\documentclass[onecolumn]{ceurart}

\usepackage{url}
\usepackage{enumitem,kantlipsum}
\usepackage{tabularx}

\usepackage{listings}

\usepackage{caption}
\begin{document}

\copyrightyear{2025}
\copyrightclause{Copyright for this paper by its authors.
  Use permitted under Creative Commons License Attribution 4.0
  International (CC BY 4.0).}

\conference{ISWC 2024 Special Session on Harmonising Generative AI and Semantic Web Technologies, November 13, 2024, Baltimore, Maryland}

\title{OAEI-LLM: A Benchmark Dataset for Understanding Large Language Model Hallucinations in Ontology Matching}

\author[1]{Zhangcheng Qiang}[
orcid=0000-0001-5977-6506,
email=qzc438@gmail.com,
]
\cormark[1]

\author[1]{Kerry Taylor}[
orcid=0000-0003-2447-1088,
email=kerry.taylor@anu.edu.au,
]

\author[2]{Weiqing Wang}[
orcid=0000-0002-9578-819X,
email=teresa.wang@monash.edu,
]

\author[1]{Jing Jiang}[
orcid=0000-0002-3035-0074,
email=jing.jiang@anu.edu.au,
]

\address[1]{Australian National University, School of Computing, 108 North Road, Acton, ACT 2601, Canberra, Australia}
\address[2]{Monash University, Faculty of Information Technology, 25 Exhibition Walk, Clayton, VIC 3800, Melbourne, Australia}

\cortext[1]{Corresponding author.}

\begin{abstract}
Hallucinations of large language models (LLMs) commonly occur in domain-specific downstream tasks, with no exception in ontology matching (OM). The prevalence of using LLMs for OM raises the need for benchmarks to better understand LLM hallucinations. The OAEI-LLM dataset is an extended version of the Ontology Alignment Evaluation Initiative (OAEI) datasets that evaluate LLM-specific hallucinations in OM tasks. We outline the methodology used in dataset construction and schema extension, and provide examples of potential use cases.
\end{abstract}


\maketitle

\section{Motivation}

Large language models (LLMs) have shown incredible capabilities in natural language generation (NLG) and question answering (QA). In the Semantic Web community, LLMs have recently been used in ontology matching (OM). Although LLMs offer a strong background knowledge base for OM, they may miss some true mappings and generate a number of false mappings. This phenomenon has been observed in several recent papers in~\cite{he2023exploring,norouzi2023conversational,hertling2023olala,qiang2023agent,giglou2024llms4om,amini2024towards}. It is mainly caused by LLM hallucinations, where LLMs tend to generate synthesised answers when they do not have sufficient background knowledge or have biased domain knowledge~\cite{zhang2023siren}.

The conference track is one of the Ontology Alignment Evaluation Initiative (OAEI)~\cite{oaei2024} campaigns that evaluates OM tasks in conference organisation. It includes several reference alignments between different ontologies related to research conference management~\cite{cheatham2014conference,solimando2014detecting,solimando2017minimizing,zamazal2017ten}. For example, \textit{``http://cmt\#Chairman''} and \textit{``http://conference\#Chair''} form a pair of valid match in the CMT-Conference alignment. LLMs can experience hallucinations when they mistakenly categorise a pair of matched entities to be unmatched entities. This often happens because it is very difficult for LLMs to understand polysemous words, such as where the word ``chair'' refers to the person who chairs the conference, and not to the furniture used at the conference. LLM hallucinations can also occur when one of the matched entities is mismatched with other entities. For example, LLMs may consider ``chair'' in the context of a research conference to have a similar meaning to ``conference chair''. Therefore, \textit{``http://conference\#Chair''} can be mismatched to \textit{``http://cmt\#ConferenceChair''}. Both the former case of missing true mappings and the latter case of generating false mappings will have a negative impact on the matching performance. It seems feasible to mitigate LLM hallucinations using fine-tuning, but there are few datasets that can be used to provide baseline references.

Established in 2004, OAEI has become the largest community of specified ontology matching challenges. Over the years, OAEI has developed several multi-ontology tracks across a wide range of topic domains. We observe that no tracks have covered the topic of LLM hallucinations in OM tasks. For this reason, we propose to extend the current OAEI datasets to construct a new dataset called OAEI-LLM, which serves to measure the degree of hallucination by LLMs in OM tasks. The new dataset compares the original human-labelled results with LLM-generated results, classifies different types of hallucinations made by different LLMs, and records the information with a new schema extension. We believe this dataset would benefit research in understanding LLM hallucinations in matching tasks and thereby research in improving LLM-driven OM.

\section{Methodology}

\subsection{Dataset Construction}

\begin{figure}[htbp]
\centering
\includegraphics[width=0.5\linewidth]{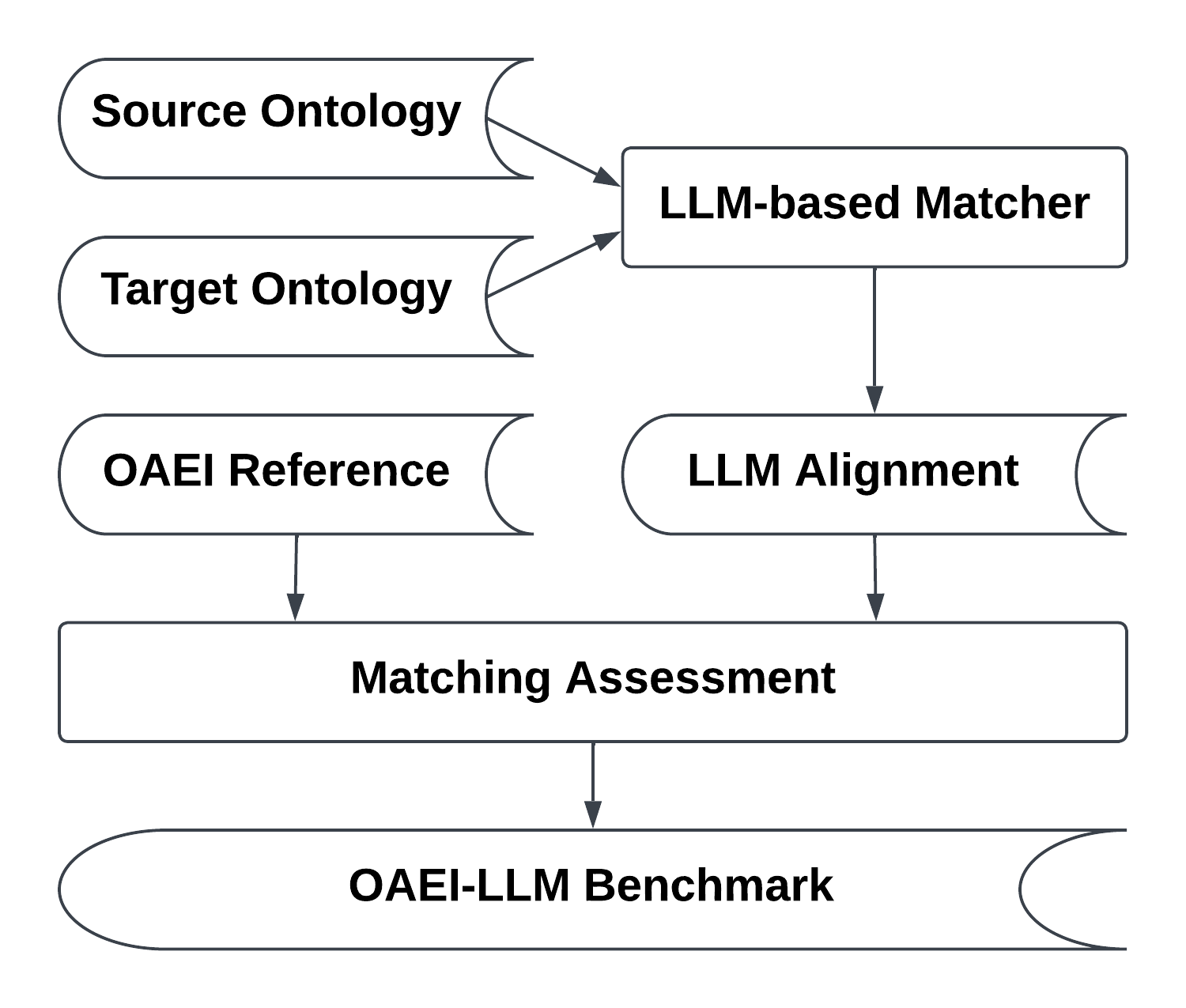}
\caption{Procedure for constructing the dataset: an LLM Alignment of the source and target ontologies is compared with the OAEI Reference and recorded in the OAEI-LLM Benchmark.}
\label{fig: methodology}
\end{figure}

\autoref{fig: methodology} illustrates the procedure for constructing the dataset. The original OAEI datasets provide three files: the source ontology ($O_s$), the target ontology ($O_t$), and their OAEI Reference ($R_{oaei}$). We implement an LLM-based OM system to generate the alignment file LLM Alignment ($A_{llm}$). The system takes $O_s$ and $O_t$ as inputs and generates a set of predicted mappings. We constrain our mappings to the case of only one-to-one mappings, that is, every entity in the source ontology is mapped to at most one entity in the target, and each entity in the target ontology occurs in at most one mapping. The matching assessment procedure detailed below will compare $R_{oaei}$ with $A_{llm}$ and find their differences.

For each mapping $(e1,e2)\in R_{oaei}$, we consider that it is not an LLM hallucination if there exists $(e1', e2') \in A_{llm}$ such that $e1 = e1'$ and $e2 = e2'$. We categorise LLM mapping errors in three ways, as shown in Table~\ref{tab: types}.

\begin{table}[htbp]
\caption{Different types of LLM mapping errors.}
\label{tab: types}
\begin{tabularx}{\textwidth}{lX}
\toprule
Type & Definition\\
\midrule
\textbf{Missing from LLM} & For a specific pair $(e1,e2) \in R_{oaei}$, there is no pair $(e1',e2') \in A_{llm}$ that satisfies $(e1 = e1')$ exclusive-or $(e2 = e2')$. \\
\textbf{Missing from OAEI} & For a specific pair $(e1',e2') \in A_{llm}$,  there is no pair $(e1,e2) \in R_{oaei}$ that satisfies $(e1 = e1')$ exclusive-or $(e2 = e2')$.\\
\textbf{Incorrect} & For a specific pair $(e1,e2) \in R_{oaei}$, there exists a pair $(e1',e2') \in  A_{llm}$ that either satisfies $e1 = e1'$ and $e2 \neq e2'$ or satisfies $e1 \neq e1'$ and $e2 = e2'$.\\
\bottomrule
\end{tabularx}
\end{table}

The matching assessment process identifies the alignment from which a missing mapping is missing. It also determines a categorisation of incorrect mappings as follows:

\begin{itemize}[wide, noitemsep, topsep=0pt, labelindent=0pt]
\item False-mapping: LLMs map the entity to an irrelevant entity.
\item Disputed-mapping: LLMs map the entity to a relevant entity, but it does not align with the OAEI Reference. We call this mapping type ``disputed'' because we cannot guarantee the OAEI Reference is always correct. LLMs may discover a more precise matching entity.
\item Align-up: LLMs map the entity to its intended superclass.
\item Align-down: LLMs map the entity into its intended subclass.
\end{itemize}

In practice, the notions of \emph{relevant} and \emph{intended} are implemented by the chosen matching assessment process. We use an LLM-based evaluator for matching assessment, and directly query the LLM for relevance and superclass and subclass relationships. At this time we cannot recommend a particular LLM to be used.

\subsection{Schema Extension}

The OAEI tracks use a general alignment format derived from the Expressive and Declarative Ontology Alignment Language (EDOAL)~\cite{edoal2011}. An example to represent a valid match of \textit{``http://cmt\#Chairman''} and \textit{``http://conference\#Chair''} is illustrated as follows: Each mapping is recorded in a tag \textit{<Cell>}. The schema uses the tags \textit{<entity1>} and \textit{<entity2>} to record the pair of entities, the tag \textit{<measure>} to present their similarity, and the tag \textit{<relation>} to indicate their relations.

We extend the current EDOAL mapping schema to record the new information related to LLM hallucinations. The Simple Standard for Sharing Ontological Mappings (SSSOM)~\cite{matentzoglu2022simple} could be an alternative base to extend for our purpose. Code Snippet~\ref{lst: example} illustrates an example of an OAEI-LLM reference alignment using our proposed extended EDOAL mapping schema. For each mapping, different types of hallucinations can occur in different LLMs. Each of them is recorded in the tag \textit{<hallucination>}. In the extended version of the example (shown in blue, green, and purple), the first segment depicts the LLM hallucinations that occurred in claude-3-sonnet~\cite{llm-claude-3-sonnet}, which is a case of a mapping missing from the LLM. The second segment illustrates the LLM hallucinations that occurred in llama-3-8b~\cite{llm-llama-3-8b}, indicating an incorrect mapping to the entity ``http://cmt\#ConferenceChair''. The matching assessment classifies this mapping as a ``disputed mapping''.

\begin{lstlisting}[label=lst: example, caption=An example of an OAEI-LLM reference alignment using our proposed extended EDOAL mapping schema. Text generated by AI tool Agent-OM is shown in green. Text generated by AI tool LLM-based evaluator is shown in purple.]
<map>
    <Cell>
        <entity1 rdf:resource="http://cmt#Chairman"/>
        <entity2 rdf:resource="http://conference#Chair"/>
        <measure rdf:datatype="xsd:float">1.0</measure>
        <relation>=</relation>
        (*@\textcolor{blue}{<hallucination>}@*)
            (*@\textcolor{blue}{<llm>claude-3-sonnet</llm>}@*)
            (*@\textcolor{blue}{<category>}\textcolor{teal}{missing}\textcolor{blue}{</category>}@*)
            (*@\textcolor{blue}{<source>}\textcolor{teal}{llm}\textcolor{blue}{</source>}@*)
        (*@\textcolor{blue}{</hallucination>}@*)
        (*@\textcolor{blue}{<hallucination>}@*)
            (*@\textcolor{blue}{<llm>llama-3-8b</llm>}@*)
            (*@\textcolor{blue}{<category>\textcolor{teal}{incorrect}</category>}@*)
            (*@\textcolor{blue}{<entity1 rdf:resource=\textcolor{teal}{"http://cmt\#ConferenceChair"}/>}@*)
            (*@\textcolor{blue}{<type>\textcolor{violet}{disputed mapping}</type>}@*)
        (*@\textcolor{blue}{</hallucination>}@*)
    </Cell>
</map>
\end{lstlisting}

\section{Potential Use Cases}

\subsection{Benchmarking LLMs for OM Tasks}

Different LLMs can perform differently on the same OM task. However, running an LLM-based matcher on a traditional OAEI dataset can only observe the differences in precision, recall, and F1 score. Using the extended version of OAEI-LLM, we can additionally quantify the LLM errors made by counting each of the different types of errors we have identified. Therefore, it becomes possible to understand the tendency of LLMs to generate incorrect answers or to provide relevant but not precise answers. This could also be a valuable supplement to enhance an LLM leaderboard for OM tasks. Indeed, this same additional information on matching failures could also be helpful if adopted more widely by non-LLM ontology matchers.

\subsection{A Dataset for Fine-tuning LLMs Used in OM Tasks}

Fine-tuning is a common approach used in LLMs to inject domain knowledge, but a significant amount of data typically needs to be provided. Unlike the traditional OAEI datasets, which only provide information on matched entities, OAEI-LLM provides extensive information identifying when and how LLM hallucinations occur, and even the mismatched entities caused by LLM hallucinations. This could be used to generate high-quality training data to fine-tune LLMs.

\section{Limitations}

In this paper, we only demonstrate the simplest one-to-one equivalent mappings. More generally, the reference alignment can be incomplete or contain one-to-many mappings, while the matching type can also go beyond equivalence to include subsumption matching and complex matching. Such sophisticated cases may cause additional types of LLM hallucinations. Some LLM hallucinations could be mitigated by providing more ontology structure as context to the LLM, for example, by employing the locality principle of traditional matchers~\cite{jimenez2011logic}. Nevertheless, the proposed benchmark dataset will be useful until LLM-based matchers achieve perfection.

In our work to date we use a moderately simple LLM-based evaluator to perform the matching assessment. The assessment may vary depending on the LLM used. A more sophisticated matching assessment could use a voting process by multiple LLMs or human expert judgement to generate a more reliable OAEI-LLM benchmark dataset.

\section{Further Work}

The OAEI tracks can be divided into three categories: schema matching (i.e. TBox matching), instance matching (i.e. ABox matching), and interactive matching (i.e. matching that involves user interaction). Correspondingly, OAEI-LLM will also have three variants.

\begin{itemize}[wide, noitemsep, topsep=0pt, labelindent=0pt]
\item OAEI-LLM-T will focus on comprehending LLM hallucinations in the TBox, with a specific interest in the formal use of terminologies in ontology design and implementation.
\item OAEI-LLM-A will focus on addressing LLM hallucinations in the ABox, in which an ontology contains a schema together with many instance entities (similar to ontology-based knowledge graphs).
\item OAEI-LLM-I will focus on understanding how user interaction could be used to mitigate LLM hallucinations and estimating the level of user interaction required for LLM-based OM tasks.
\end{itemize}

\paragraph*{Acknowledgements}
\begin{small}
The authors thank the reviewers for providing valuable feedback. The authors thank the Commonwealth Scientific and Industrial Research Organisation (CSIRO) for supporting this project. According to the OAEI data policy (retrieved December 1, 2024), ``OAEI results and datasets, are publicly available, but subject to a use policy similar to \href{https://trec.nist.gov/results.html}{the one defined by NIST for TREC}. These rules apply to anyone using these data.'' Please find more details from the official website: \url{https://oaei.ontologymatching.org/doc/oaei-deontology.2.html}.
\end{small}

\newpage
\bibliography{qiang-bibliography-iswc}

\end{document}